\documentclass[preprint,12pt]{elsarticle}          

\usepackage[utf8x]{inputenc}
\usepackage{graphicx}
\graphicspath{ {./images/} }
\usepackage{tikz}
\usetikzlibrary{positioning,shadows,shapes,arrows}
\usepackage{float}
\usepackage{multicol}
\usepackage[ruled,vlined]{algorithm2e}
\usepackage{pgfplots}
\definecolor{blue}{HTML}{1F77B4}
\definecolor{orange}{HTML}{FF7F0E}
\definecolor{green}{HTML}{2CA02C}

\begin{document}

\begin{frontmatter}

\title{Teach me to play, gamer! Imitative learning in computer games via linguistic description of complex phenomena and decision tree}

\author[a1,a2]{Clemente Rubio-Manzano}
\ead{clrubio@ubiobio.cl}
\author[a2]{Tom\'as Lermanda}
\author[a3]{Claudia Mart\'inez-Araneda}  
\author[a2]{Christian Vidal}  
\author[a2]{Alejadra Segura}  
         
\address[a1]{Department of Mathematics, University of Cadiz, Cadiz, Spain.}
\address[a2]{Department of Information Systems, University of the Bio-Bio, Concepci\'on, Chile}
\address[a3]{University of Cat\'olica Sant\'isima Concepci\'on, Chile.}

\begin{abstract}
In this article, we present a new machine learning model by imitation based on the linguistic description of complex phenomena. The idea consists of, first, capturing the behaviour of human players by creating a computational perception network based on the execution traces of the games and, second, representing it using fuzzy logic (linguistic variables and if-then rules). From this knowledge, a set of data (dataset) is automatically created to generate a learning model based on decision trees. This model will be used later to automatically control the movements of a bot. The result is an artificial agent that mimics the human player. We have implemented, tested and evaluated this technology. The results obtained are interesting and promising, showing that this method can be a good alternative to design and implement the behaviour of intelligent agents in video game development.
\end{abstract}


\end{frontmatter}

\section{Introduction}

In recent years, videogames have surpassed cinema in market share and have positioned themselves as one of the most popular and complex software systems in the world. In addition to entertainment purposes, they have been consolidated in other disciplines such as education or scientific research, where for example, \cite{gemine2012imitative} mentions that \textit{``modern video games have become an alternate, low-cost yet rich environment for assessing machine learning algorithms''}. \par

One of the most important phases in the development of video games is the modelling and programming of opponents, usually known as NPCs from their acronym in English, non-player characters. NPCs are autonomous and intelligence agents that play a fundamental role in their final quality; that is, they allow games to be made more \textit{``challenging and enjoyable''} \cite{feng2016towards}. \par

The role of NPCs in video games has at least two dimensions: quantity and quality. The quantity refers to having an adequate number of actors that provide the viewer with the feeling of participating in a real world; the second is the ability to create the illusion of credibility and is directly related to the behaviour of the actors \cite{borovikov2019towards}. Currently, NPCs are programmed using deterministic computational techniques such as fuzzy state machines (FSMs) or scripting languages. Although these techniques are easy to use and quick to implement, some limitations have been pointed out in the literature, as mentioned in \cite{melendez2009controlling}: \textit{''there are some undesirable side effects with their use such as behaviours being less believable and more predictable due to the deterministic nature of these techniques''}. \par

On the other hand, although more techniques have been incorporated into advances, such as behaviour trees (BTs) \cite{marcotte2017behavior}, the main limitation of FSMs or BTs is that their design is top down and not bottom up. Including a bottom-up phase (data-driven) can improve the FSMs and BTs and the knowledge of the designers about their game. Recently, bottom-up techniques, such as gameplay metrics and machine learning ones, have been developed \cite{melendez2009controlling}. Regarding this last point,  in the literature, some challenges and open problems related to the programming of NPCs have been raised:
\begin{enumerate}
 \item \textit{''Developing automated agents that intelligently perform complex real-world tasks is time consuming and expensive. The most expensive part of developing these intelligent task performance agents involves extracting knowledge from human experts and encoding it into a form useable by automated agents''} \cite{van1999learning};
 \item \textit{''Modern Computer Game AI still relies on rule-based approaches, so far failing to develop a convincing, human-like opponent''} \cite{thurau2004learning};
 \item \textit{“Additionally, while players get more familiar with the game mechanics and improve their skills and devise new strategies, agents do not change and eventually become obsolete"} \cite{gemine2012imitative}.
\end{enumerate}

One of the techniques used to solve these challenges has been learning by imitation, which is a very powerful learning mechanism that humans have and allows the acquisition of knowledge from observation \cite{thurau2007bayesian}. In \cite{feng2016towards}, for example, it is mentioned that \textit{''although reinforcement learning (RL) has been a promising approach to creating the behavior models of NPCs, an initial stage of exploration and low performance is typically required. On the other hand, imitative learning is an effective approach to pre-building an NPC’s behavior model by observing the opponent’s actions, but learning by imitation limits the agent’s performance in relation to that of its opponents."}

One of the requirements of this type of learning is as stated in \cite{thurau2004imitation}: \textit{"imitation requires a mechanism to map perceptions onto actions"}. This suggests that the theory of perceptions \cite{zadeh2001new}, and therefore, the management of imprecision plays an important role in achieving good models of imitation learning.

Although several learning algorithms have been applied in computer games, none have actually been used to dictate agent behaviour directly by using Linguistic Description of Complex Phenomena (LDCP). Another innovation of this paper is that LDCP is typically used to analyse a report in natural language which is automatically genereted from the data; however, in this work, LDCP is used to generate a dataset that will be used for training a bot.

More precisely, it is intended to create a general method that allows it to map computational perceptions onto moves for training a bot from the human play sessions. To do this, the behaviour of the players will first be captured from a computational perception network that will be created from the observations (execution traces) of the games. Later, we will use the computational perception network to create a dataset that can be used to generate a decision tree. A decision tree will be incorporated into the video game so that through a classification process, the movements of the bots can be controlled automatically. Our results highlight that our model produces agents that behave like human players. As our model is based on the theory of computational perceptions, it has a high level of interpretability and makes the decision-making process of the bots transparent. At the same time, it provides solutions to the previously mentioned open problems since it allows for obtaining expert knowledge of the players' games in an automatic way that can help developers when programming bots. Both the system and the demonstration videos that show the results can be publicly accessed. The development can be found at the following URL:
\begin{center}
\verb+https://github.com/pp1856/Agent-Training+
\end{center}

The following are demonstration videos of bot training using LDCP and decision trees:

\begin{itemize}
\item Test 1 (7 attributes): \\ \verb+https://www.youtube.com/watch?v=A8Y4by5Llng+
\item Test 2 (9 attributes): \\ \verb+https://www.youtube.com/watch?v=CCZt5IHKjos+
\item Test 2 (17 attributes): \\ \verb+https://www.youtube.com/watch?v=KfZpOQ6Qvsg+
\item Test 3 (9 attributes): \\ \verb+https://www.youtube.com/watch?v=Gx-x9lbby30+
\item Test 3 (17 attributes): \\ \verb+https://www.youtube.com/watch?v=87Rlh7GP9PE+
\end{itemize}

The rest of the article is organized as follows. In Section 2, the basic concepts necessary to understand the rest of the work will be described. In Section 3, we explain in detail the model of learning by imitation based on LDCP. Section 4 explains the tracking and attention phases and how the data is captured and prepared for the perception phase that is described in Section 5, where the computational perception network that is responsible for transforming the data will be presented with data on linguistic knowledge. The training and classification process using decision trees from previous knowledge is given in Section 6. The evaluation of the model is performed in Section 7, and the conclusions are presented in Section 8.

\section{Preliminary concepts}
\label{preliminary-concepts}

This section introduces some needed definitions to understand the rest of the paper.

\subsection{Our Computer Game}

A computer game is an interactive system that provides players with immersive and pleasurable experiences. This is formed by three components: (i) the input-output devices that provide players with controllers and interfaces in order for players to interact with the game; (ii) the objectives and the rules of the game that establish the rules of interaction among the different elements; and (iii) the program that describes how the game is implemented at the code level. The game world is the environment in which the game takes place. It is formed by either static or active elements. The actors are the entities that exist within the game world and that can interact with the elements of the game world and with other actors. The mechanism of interaction is fixed by means of the rules of the game. The player is an actor whose movements are performed by human players. The agents are actors whose movements are performed by a virtual player, i.e., specialized software.

\begin{figure}
\centering
\includegraphics[height=5.0cm]{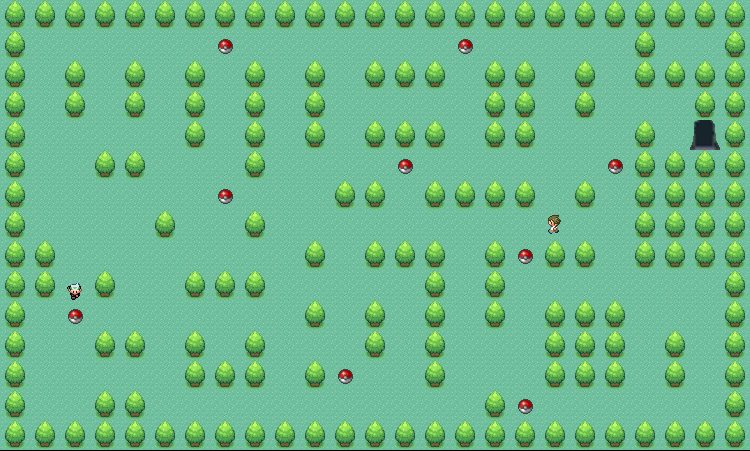}
\caption{2D computer game developed by the research}
\label{fig-computergame}
\end{figure}

We employ our own 2D computer game based on the pac-man game created for this reseach. Others versions of this computer game has been previusly used in previous works \cite{RUBIOMANZANO201627,rubio2019fuzzy}. In each play session, the player navigates through a maze containing rewards and one opponent. The goal of the game is to collect all the rewards and escaping from the opponent. The opponent roams the scene and chase the player. The game ends when all rewards has been collected or when the player is captuted (see Figure~\ref{fig-computergame}). 

\subsection{Linguistic Description of Complex Phenomena}

Automatic Linguistic Description of Complex Phenomena aims to extract and represent knowledge by using natural language sentences (report) as if they were produced by a human expert, describing the most relevant aspects of a phenomenon for certain users in specific contexts.

LDCP is based on the computational theory of perceptions \cite{zadeh2001new}. It is grounded in the fact that human cognition is based on the role of perceptions and their remarkable capability to granulate information to perform physical and mental tasks without any traditional measurements and computation. LDCP has been used in previous works to generate linguistic advice for saving energy at home \cite{conde2018toward}.

The architecture is based on the concept of Computational Perception (CP). A CP is a pair (A,W) described as follows:

\begin{description}
\item $A = (u_1,\ldots,u_n)$ is a vector of linguistic expressions (words or sentences in Natural Language) that represents
      the whole linguistic domain of CP. 
\item $W = (w_1,\ldots,w_n)$ is a vector of the validity degrees $w_i \in [0,1]$ of each $u_i$. $w_i$ represents the suitability of $u_i$ to describe the current perception of a specific aspect of the monitored phenomenon.
\end{description}

For example:

\begin{description}
\item $u_1$=``The current situation is dangerous'', $w_1=0.8$
\item $u_2$=``The current situation is risky'', $w_2=0.2$
\item $u_3$=``The current situation is easy'', $w_3=0.0$
\item $u_4$=``The current situation is safe'', $w_4=0.0$
\end{description}

We use Perception Mappings (PM) to aggregate CPs. We distinguish two kind of PMs, namely, First Order PMs (1PMs) and Second Order PMs (2PMs). We define a 1PM as a triple (Z,y,$g$); where Z is a special type of CP with a numerical value $z$, y is a 1CP, $g$ is a function $W=g(z)$, the function $g(z)$ can be implemented by using membership functions $w_i=\mu_{i}(Z)$ associated with each component $a_i$ of A and therefore: W=($\mu_{1}(Z),\mu_{2}(Z),\ldots,\mu_{n}(Z)$) where $n$ es the elements of elements in A. A 2PM is a tuple (U,y,$g$); where U is a vector of input CPs, y is the output CP and $g$ is an aggregation function implemented by using a set of fuzzy rules (see several examples of PM later on). 

\subsection{Decision Tree Classifier}

Classification using decision trees solves a classification problem by decomposing a complex decision into a set of simpler decisions. A decision tree is built from a training set, which is made up of objects. Each object is completely described by a set of attributes and a class label \cite{Murthy1998}. The attributes of an object can have numerical or nominal values ​​(unordered data). The class label is a nominal variable.

The structure of a decision tree follows a hierarchical order from top to bottom. It can be represented as a directed tree. A decision tree is made up of a root node, zero or more internal nodes, and at least one leaf node. Each node has an associated attribute, and each division of a node evaluates the value of an expression of the attribute associated with it. The attributes present in a tree are those considered good at discriminating the input information. Leaf nodes differ by having a class label associated with them.

We build our trees using the C4.5 Decision Tree Induction Algorithm. This algorithm uses the divide-and-conquer approach to construct the decision tree. This strategy, sometimes called top-down induction of decision trees, can be recursively expressed as follows:

\newenvironment{blockquote}{%
  \par%
  \medskip
  \leftskip=4em\rightskip=2em%
  \noindent\ignorespaces}{%
  \par\medskip}
\begin{blockquote}
First, select an attribute to place at the root node and make one branch for each possible value. This splits up the example set into subsets, one for every value of the attribute. Now, the process can be repeated recursively for each branch, using only those instances that actually reach the branch. If at any time all instances at a node have the same classification, stop developing that part of the tree \cite{Witten2011}. 
\end{blockquote}

The attribute to select is the one that produces the smallest tree, and the criterion used by C4.5 is the information gain. The information gain represents the informational value of creating a branch with a certain attribute. The attribute with the highest information gain value is chosen.

\begin{figure}[htb!]
\begin{center}
\begin{tikzpicture}[
    metr/.style={rectangle, draw=black, fill=white, 
        text centered, anchor=north, text=black,
        node distance=1.75cm and -0.5cm},
    resu/.style={rectangle, draw=white, fill=white, 
        text centered, anchor=north, text=black,
        node distance=1.62cm and 1.5cm},
    leaf/.style={ellipse, draw=black, fill=white, 
        text centered, anchor=north, text=black,
        node distance=1.75cm and 0.75cm},
    level distance=0.5cm, growth parent anchor=south
]
\draw[thick,dashdotted] (-3.5,-0.37) -- (3.5,-0.37);
\draw[thick,dotted] (-2.5,-2.67) -- (3.5,-2.67);
\draw[thick,loosely dotted] (-3.5,-5.1) -- (3.5,-5.1);

\node[leaf][minimum width=1cm, minimum height=0.75cm] (a1) {\scriptsize attribute1};
\node[leaf][minimum width=1cm, minimum height=0.75cm] (a2) [below left=of a1] {\scriptsize attribute2};
\node[leaf][minimum width=1cm, minimum height=0.75cm] (a3) [below right=of a1] {\scriptsize attribute3};
\node[metr][minimum width=1cm, minimum height=0.75cm] (c1) [below left=of a2] {\scriptsize class\_label1};
\node[metr][minimum width=1cm, minimum height=0.75cm] (c2) [below right=of a2] {\scriptsize class\_label2};
\node[metr][minimum width=1cm, minimum height=0.75cm] (c3) [below left=of a3] {\scriptsize class\_label1};
\node[metr][minimum width=1cm, minimum height=0.75cm] (c4) [below right=of a3] {\scriptsize class\_label2};
\node[resu][minimum width=1cm, minimum height=0.75cm] (l1) [left=of a1] {\scriptsize root node};

\draw[->] (a1) -- (a2) node[midway,above left] {a1\_value1};
\draw[->] (a1) -- (a3) node[midway,above right] {a1\_value2};
\draw[->] (a2) -- (c1) node[midway,above left] {a2\_value1};
\draw[->] (a2) -- (c2) node[midway,above right] {a2\_value2};
\draw[->] (a3) -- (c3) node[midway,above left] {a3\_value1};
\draw[->] (a3) -- (c4) node[midway,above right] {a3\_value2};

\end{tikzpicture}
\end{center}
\caption{Decision Tree structure example.}
\end{figure}
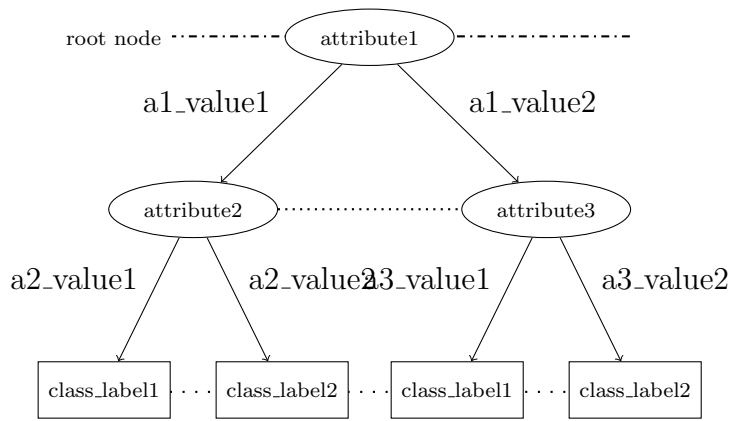

Once a tree has been built, the classification is performed with an object (also called an instance) as input data, and the tree is followed using the values ​​of the attributes of the instance until the classified label for the instance is obtained. The trees built are usually evaluated based on statistics obtained starting from the number of instances correctly and incorrectly classified.

\section{Imitation model based on linguistic description of complex phenomena}

\begin{figure*}
\centering
\includegraphics[height=6.0cm]{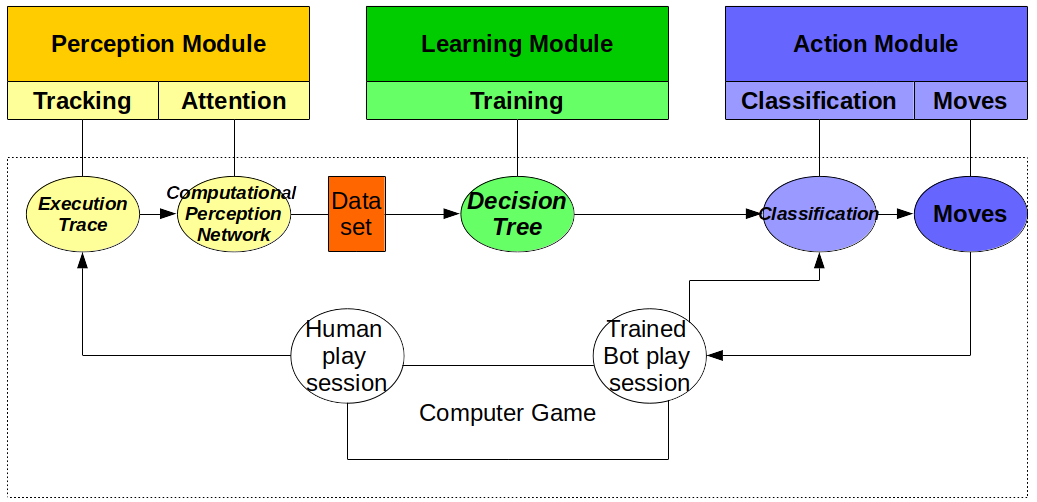}
\caption{Our imitation model based on linguistic modelling of complex phenomenon}
\label{fig-model}
\end{figure*}

In this section, we will describe in detail our model of learning by imitation based on LDCP. The model is explained as a fuzzy adaptation of the model presented in \cite{fod2002automated}. Our model is made up of five submodules: tracking, attention, training, classification, and moves. At the same time, these modules are grouped into three modules: perception, learning, and action. In Fig~\ref{fig-model}, each of the modules is shown together with their dependencies. The components in charge of carrying out the functionality of each submodule is shown (see section~\ref{perception}):
\begin{enumerate}
 \item  Tracking - Execution Traces
 \item  Attention - Computational Perception Network
 \item Training - Decision Tree
 \item Classification - Static Tree
 \item Moves - Bot
\end{enumerate}

\subsection{Perception}
\label{perception}

The perception module is made up of two submodules, tracking and attention, which are used to acquire and prepare the data so that they can be processed by a computational perception network. In our model, tracking is responsible for extracting the information from the current state of the game. More formally, at a given time $i$, the state of a game can be defined by a vector that will contain the information about what is happening in the virtual world at that moment. We call this vector the vector of the virtual world $w_i \in W$, with W being the game space where every state can be represented. The observation space $\cal O$ represents the observable information for a player and is therefore a subspace of W. We define an observation/instance as a vector $o \in \cal O$ with $O \subset W$. Considering the entire observation space, we will focus on those observations that can lead to some behaviours that cause a movement S. We define the state as a vector $s = (s_1 , s_2 , \ldots, s_n )$ as the projection of o in S.

The attention module is responsible for selecting the relevant information from the data captured in the tracking with the aim of simplifying the classification and learning. The selection is made considering the variables that intervene in the behaviour and that are believed to be key to implementing the imitation process. In particular, we will use fuzzy logic and computational perception networks (if-then rules with linguistic variables). We start from a state vector $s = (s_1 , s_2 , \ldots, s_n )$ formed by a set of precise variables. As some observations may be imprecise and depend on each player, we will handle imprecision using computational perception vectors. From a precise vector, a vector of computational perceptions can be created: $l = (l_1 , l_2 , \ldots, l_n )$. For each $s_i$, there may be $m$ perceptions $(l_1, l_2, \ldots, l_m)$ that will depend on the fuzzy linguistic definition of each $s_i$. We employ the concept of CP. A CP state is a pair (U,W) described as follows:
\begin{description}
\item $L = (l_1,\ldots,l_n)$ is a linguistic vector that represents the whole linguistic
domain of CP, which is defined from $(s_1 , s_2 , \ldots, s_n )$.
\item $W = (w_1,\ldots,w_n)$ is a vector of the validity degrees $w_i \in [0,1]$ of each $l_i$. $w_i$ represents the suitability of $l_i$ to describe the current perception of a specific aspect of the monitored phenomenon.
\end{description}

Recapitulating, from a set of data obtained from an execution trace, we will obtain a set of observations or precise instances that can be transformed into a set of computational perceptions (list of pairs: (linguistic value, degree of validity)). An implementation of this module is explained in detail in section~\ref{perceptions}.

\subsection{Learning}

\begin{figure*}
\begin{center}
\begin{tikzpicture}[
    lab1/.style={rectangle, draw=black, fill=white, 
        text centered, anchor=north, text=black,
        node distance=1.75cm and 1.5cm},
    lab2/.style={rectangle, draw=black, fill=white, 
        text centered, anchor=north, text=black,
        node distance=1.75cm and 0.10cm},
    lab3/.style={rectangle, draw=black, fill=white, 
        text centered, anchor=north, text=black,
        node distance=1.75cm and 0.25cm},
    leaf/.style={ellipse, draw=black, fill=white, 
        text centered, anchor=north, text=black,
        node distance=1.75cm and 1.5cm},
    level distance=0.5cm, growth parent anchor=south
]

\node[leaf][minimum width=1cm, minimum height=0.75cm] (a1) {\tiny attitude};
\node[leaf][minimum width=1cm, minimum height=0.75cm] (a2) [below left=of a1] {\tiny movement};
\node[lab1][minimum width=1cm, minimum height=0.75cm] (c1) [right=of a2] {\tiny get away};
\node[leaf][minimum width=1cm, minimum height=0.75cm] (a3) [below right=of a1] {\tiny movement};
\node[lab1][minimum width=1cm, minimum height=0.75cm] (c2) [right=of a3] {\tiny get away};
\node[lab2][minimum width=1cm, minimum height=0.75cm, align=center, font=\tiny\linespread{0.8}\selectfont] (c4) [below left=of a2] {go ahead};
\node[lab3][minimum width=1cm, minimum height=0.75cm, align=center, font=\tiny\linespread{0.8}\selectfont] (c3) [left=of c4] {go ahead};
\node[lab2][minimum width=1cm, minimum height=0.75cm, align=center, font=\tiny\linespread{0.8}\selectfont] (c6) [below right=of a2] {go ahead};
\node[lab3][minimum width=1cm, minimum height=0.75cm, align=center, font=\tiny\linespread{0.8}\selectfont] (c5) [left=of c6] {get away};
\node[lab2][minimum width=1cm, minimum height=0.75cm, align=center, font=\tiny\linespread{0.8}\selectfont] (c7) [below left=of a3] {get away};
\node[lab3][minimum width=1cm, minimum height=0.75cm, align=center, font=\tiny\linespread{0.8}\selectfont] (c8) [right=of c7] {go ahead};
\node[lab2][minimum width=1cm, minimum height=0.75cm, align=center, font=\tiny\linespread{0.8}\selectfont] (c9) [below right=of a3] {get away};
\node[lab3][minimum width=1cm, minimum height=0.75cm, align=center, font=\tiny\linespread{0.8}\selectfont] (c10) [right=of c9] {get away};

\draw[->] (a1) -- (a2) node[midway, left] {\tiny wise};
\draw[->] (a1) -- (c1) node[midway, left] {\tiny brave};
\draw[->] (a1) -- (a3) node[midway,right] {\tiny prudent};
\draw[->] (a1) -- (c2) node[midway,right] {\tiny passive};

\draw[->] (a2) -- (c3) node[midway, left] {\tiny good};
\draw[->] (a2) -- (c4) node[midway,below left] {\tiny scared};
\draw[->] (a2) -- (c5) node[midway,below] {\tiny kamikaze};
\draw[->] (a2) -- (c6) node[midway,below right] {\tiny bad};

\draw[->] (a3) -- (c7) node[midway,below left] {\tiny good};
\draw[->] (a3) -- (c8) node[midway,below] {\tiny scared};
\draw[->] (a3) -- (c9) node[midway,below right] {\tiny kamikaze};
\draw[->] (a3) -- (c10) node[midway,right] {\tiny bad};
\end{tikzpicture}
\end{center}
\caption{Decision tree generated with C4.5}
\label{decision}
\end{figure*}
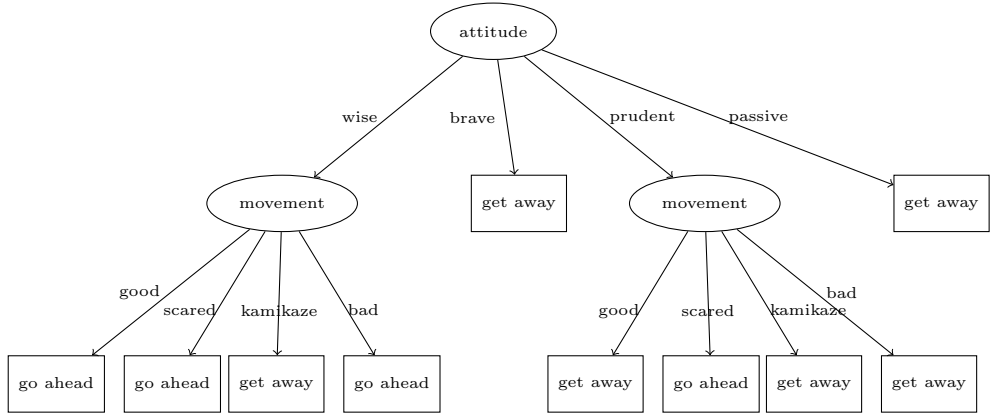

For the learning process, we will assume that the set of games (traces) is available and that they are generated from expert players. Each game contains the vectors presented in the previous phase. We are not going to make assumptions about the quality of the dataset. The objective is, therefore, to learn a relationship of the type P: CPs $\leftarrow$ M that, based on perceptions, gives us the best move to perform given the state of the game at a given moment. The learning result (decision tree) can be seen in Figure~\ref{decision} and the process is explained in detail in section~\ref{classification}.

\subsection{Classification and Movements}

The classification and movements submodule is in charge of carrying out the imitation process through the execution of the movements acquired in the learning phase. More precisely, the process is as follows: a row is read from the execution trace, the data are converted to CPs, which are organized in an instance, the instance is classified and the label resulting from the classification is used as input data in a function of search whose result determines the direction in which the agent moves. Once the direction is determined, the corresponding movement is carried out. The algorithms that perform these tasks are shown in Algorithm~\ref{a1}, ~\ref{a2} y ~\ref{a3}
   
\begin{algorithm}
\SetAlgoLined
\KwIn{$T$ (execution trace)}
\KwOut{void (executes a movement $M_i$)}

R = read last row(T) \;
I = create instance(R) \;
V = classify(I) \;
R = search (V) \;

\ForEach{$M_i$: possible movements}{
 
  \uIf{R==$M_i$}{
     executes movement $M_i$\;
   }
}

\caption{function toMove}
\label{a1}
\end{algorithm}

To generate the decision trees, we use J48, which is the implementation of the C4.5 algorithm from the Weka library. The resulting trees are exported as a static classifier (Java class), which provides us with a convenient function to perform the classification. The classification function is where an instance is used to create the data in the table in real time to determine the class label, the variable that determines the final decision to be made.

\begin{algorithm}
\SetAlgoLined
\KwIn{Instance $I$ to be classified}
\KwOut{Class $P$ most likely for the instance $I$}
 
P = NaN\;
P = nX(I)\;

\Return{P}\;
\caption{function classify}
\label{a2}
\end{algorithm}

The function nX is a different function for each node of the tree, and X represents a hexadecimal number assigned to each node. The function is called in principle from the root node, and each subsequent call is made from an internal node.

\begin{algorithm}
\SetAlgoLined
\KwIn{Instance I to be classified}
\KwOut{Predicted most likely class P for the instance I}
 
(Note I is represented as an array of capacity C and P is represented as a double value.)

P = NaN \;

\ForEach{decision rule R}{
  \eIf{I[s] is compared with R evaluates to true}{
     either P = nX(I) if there are more childnodes OR 
            P = D if there are none
   }

}

(Note s is an index of the instance evaluated in a specific node and D is a value of the class)\;

\Return{P}\;
\caption{function nX}
\label{a3}
\end{algorithm}

In each variation of the function, a different test is performed, and the different values ​​of an attribute associated with a node are evaluated. It follows from node to node until obtaining the label (P) resulting from the classification.

\section{Tracking}
\label{tracking}

For tracking, we will use the concept of execution trace, which is a technique used to record the relevant information that occurs during the execution of a program. Execution traces are commonly used as a performance analysis and debugging tool. In our case, due to the nature of video games, we will use them to capture and store the data generated from the movements made by human players, agents and opponents. In this way, the traces perform the observation function (key in frameworks based on learning by imitation).

In particular, the data captured correspond to values ​​associated with the state of the game at the time the movement is made. Specifically, the locations of the relevant entities include the player or agent, the adversary, and the reward closest to the player or agent. The time of the game, the event of the capture of a reward and the entity that performs the action are also recorded.

From the captured data, metrics are defined that provide useful data that cannot be inferred using the values ​​captured individually. Using metrics, the state of the game and the behaviour shown by the player or agent can be interpreted. In this case, the defined metrics are as follows.
\begin{itemize}
 \item \textbf{Protection.} Percentage of obstacles present between the player or agent and the opponent; a rectangular area created from the location of the player or agent and the opponent is used.
 \item \textbf{Distance.} Distance between two entities E1 and E2.
 \item \textbf{Time.} Time elapsed from the start of the game until the moment the action was performed.
\end{itemize}

\section{Attention}
\label{perceptions}

Attention is implemented using the concept of a computational perception network. We will use a modified version of the network of perceptions presented in \cite{RUBIOMANZANO201627}; in this case, the computational perceptions (CPs) have been reduced. Although all first-order computational perceptions are maintained (1CP), two second-order computational perceptions (2CP), CP Ability and CP Skill, are eliminated. This is because our objective is not to evaluate the performance of the player but rather to use the perceptions that best represent the actions of a player. In this way, the 2 CPs used correspond to CP Situation, CP Attitude and CP Movement. The metrics, linguistic variables and rules used have been redefined for this development, so they will be explained in detail below.

\subsection{First-order computational perceptions}

We use numerical data obtained from the variables and metrics previously defined as input data for the membership functions that will determine the degrees of validity of each 1CP.

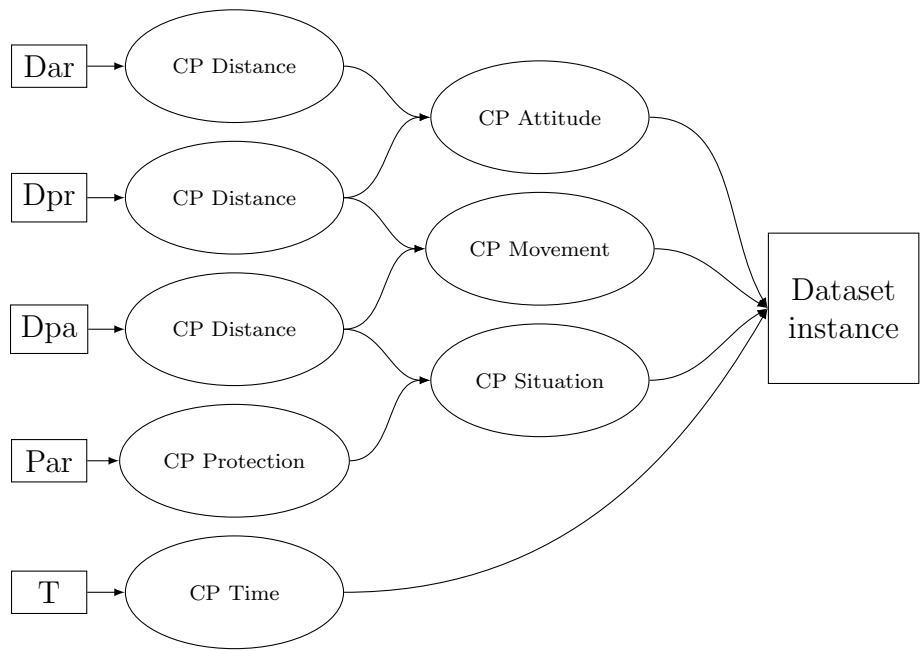
\begin{figure*}
\begin{center}
\begin{tikzpicture}[
    metr/.style={rectangle, draw=black, fill=white, 
        text centered, anchor=north, text=black,
        node distance=1.75cm and 0.5cm},
    resu/.style={rectangle, draw=black, fill=white, 
        text centered, anchor=north, text=black,
        node distance=1cm and 2cm},
    leaf/.style={ellipse, draw=black, fill=white, 
        text centered, anchor=north, text=black,
        node distance=1.75cm and 2cm},
    level distance=0.5cm, growth parent anchor=south
]

\node[leaf][minimum width=2.9cm, minimum height=1.5cm] (1CP1) {\scriptsize CP Distance};
\node[leaf][minimum width=2.9cm, minimum height=1.5cm] (1CP2) [below of=1CP1] {\scriptsize CP Distance};
\node[leaf][minimum width=2.9cm, minimum height=1.5cm] (1CP3) [below of=1CP2] {\scriptsize CP Distance};
\node[leaf][minimum width=2.9cm, minimum height=1.5cm] (1CP4) [below of=1CP3] {\scriptsize CP Protection};
\node[leaf][minimum width=2.9cm, minimum height=1.5cm] (1CP5) [below of=1CP4] {\scriptsize CP Time};
\node[leaf][minimum width=2.9cm, minimum height=1.5cm]  (2CP1) [above right=of 1CP3] {\scriptsize CP Attitude};
\node[leaf][minimum width=2.9cm, minimum height=1.5cm] (2CP2) [below of=2CP1] {\scriptsize CP Movement};
\node[leaf][minimum width=2.9cm, minimum height=1.5cm] (2CP3) [below of=2CP2] {\scriptsize CP Situation};
\node[resu][minimum width=2cm, minimum height=2cm, align=center] (outp) [below right=of 2CP1] {Dataset \\ instance};
\node[metr][minimum width=1cm] (M1) [left=of 1CP1] {Dar};
\node[metr][minimum width=1cm] (M2) [below of=M1] {Dpr};
\node[metr][minimum width=1cm] (M3) [below of=M2] {Dpa};
\node[metr][minimum width=1cm] (M4) [below of=M3] {Par};
\node[metr][minimum width=1cm] (M5) [below of=M4] {T};

\draw[-latex] (M1.east) to[out=0,in=-180] (1CP1.west);
\draw[-latex] (M2.east) to[out=0,in=-180] (1CP2.west);
\draw[-latex] (M3.east) to[out=0,in=-180] (1CP3.west);
\draw[-latex] (M4.east) to[out=0,in=-180] (1CP4.west);
\draw[-latex] (M5.east) to[out=0,in=-180] (1CP5.west);
\draw[-latex] (1CP1.east) to[out=0,in=-180] (2CP1.west);
\draw[-latex] (1CP2.east) to[out=0,in=-180] (2CP1.west);
\draw[-latex] (1CP2.east) to[out=0,in=-180] (2CP2.west);
\draw[-latex] (1CP3.east) to[out=0,in=-180] (2CP2.west);
\draw[-latex] (1CP3.east) to[out=0,in=-180] (2CP3.west);
\draw[-latex] (1CP4.east) to[out=0,in=-180] (2CP3.west);
\draw[-latex] (1CP5.east) to[out=0,in=240] (outp.west);
\draw[-latex] (2CP1.east) to[out=0,in=120] (outp.west);
\draw[-latex] (2CP2.east) to[out=0,in=150] (outp.west);
\draw[-latex] (2CP3.east) to[out=0,in=210] (outp.west);
        
\end{tikzpicture}
\end{center}
\caption{Computational perception network.}
\end{figure*}

\begin{itemize}
\setlength\itemsep{1em}
\item \textbf{CP Distance.} This is the perception used to perceive the space between entities. Given a scenario divided into cells of equal size and considering that an entity occupies the space of a cell, the distance is calculated according to the number of cells between a pair of entities. If each cell is represented by a location x, y, then the distance D between two locations is calculated using the following equation: D = (x-x’) + (y-y’). We define three different distance CPs: the distance between the player and the opponent, the distance between the player and the reward closest to the player, and the distance between the opponent and the reward closest to the player. We denounce this 1CP as follows:

$Z = [0,N]$, where N is the maximum distance in the game scenario. 

$A = (small; medium; large)$

g: This function is constructed using three labels that are represented with trapezoidal membership functions: small (0,0,2,6), medium (2,6,6,10), and large (6,10, N, N).

\item \textbf{CP Protection.} This is the perception used to perceive how obstructed the area is between the player and the opponent. Given a split stage in cells of equal size and considering that an entity occupies the space of a cell, the protection is calculated by determining the percentage of obstacles present in the area between the player and the opponent. If each cell is represented by a location x, y, the rectangular area formed by the player and opponent locations is calculated, then we count the number of obstacles present (Opr) and possible (Opo) in the area. Finally, we divide them to obtain the percentage P of obstacles present. The calculation reduces to the following equation: P = Opr/Opo. We define this 1CP as follows: 

$Z = [0,1]$
$A = (low; medium; high)$
g: This function is constructed using three labels that are represented with trapezoidal membership functions: low (0,0,0.25,0.41), medium (0.25,0.41,0.58,0.75), and high (0.58,0.75,1,1).

\item \textbf{CP Time.} This is the perception used to perceive the time elapsed since the start of the game. Time is calculated by determining the amount of time in milliseconds elapsed from the start of the game until the moment a move was made. Given the current time t and the time at the start of game t’, the calculation of the difference results in the elapsed time T. The calculation reduces to the following equation: T = t - t’. We define this 1CP as follows:

$Z = [0,N]$, N being the maximum time established for the duration of a game. 

$A = (little; reasonable; a lot)$ 

g: This function is constructed using three labels that are represented with trapezoidal membership functions: 

$low (0,0,9K,15K)$; \\
$medium (9K,15K,21K,27K)$ and; \\
$high (21K,27K, N, N)$\\
(K is represented *1000 e.g. 9K is 9000)
\end{itemize}

\subsection{Second-Order Computational Perceptions}

\begin{itemize}
    \setlength\itemsep{1em}
    \item \textbf{CP Situation.} This is the perception used to perceive the state in which the player is in relation to the opponent. Four possible situations are defined: risky, dangerous, safe, and easy. The situation is determined by evaluating the 1CP Protection and Distance (player-opponent). With the values ​​of the linguistic variables obtained from these 1CPs, in this case, protection (low, medium, high) and distance (small, medium, large), an if-then rule set is used to determine the value of the CP. The rules used are the following: {risky $\leftarrow$medium, small; risky $\leftarrow$large, small; dangerous $\leftarrow$ low, small; safe $\leftarrow$ medium, medium; safe $\leftarrow$ high, medium; easy $\leftarrow$ low, medium; easy $\leftarrow$ low, high; safe in any other case}.
    \item \textbf{CP Attitude.} This is the perception used to perceive the actions of the player. Four possible attitudes are defined: wise, brave, prudent and passive. The attitude is determined by evaluating the 1CP Distance (player-reward) and Distance (opponent-reward). With the values ​​of the linguistic variables obtained from this 1CP, in this case, Distance (small, medium, large), a set of if-then rules is used to determine the value of the CP. The rules used are the following: {wise $\leftarrow$ small, medium; prudent $\leftarrow$small, large; prudent $\leftarrow$ medium, large; brave $\leftarrow$ small, small; brave $\leftarrow$ medium, medium; wise $\leftarrow$ medium, small; wise $\leftarrow$ large, small; passive $\leftarrow$large, large; passive in any other case}.
    \item \textbf{CP Movement.} This is the perception used to perceive the type of movement made by the player. Four possible moves are defined: good, scared, kamikaze and bad. Movement is determined by evaluating the 1CP Distance (player-reward) and Distance (player-opponent). With the values ​​of the linguistic variables obtained from this 1CP, in this case, Distance (small, medium, large), a set of if-then rules is used to determine the value of the CP. The rules used are the following: {good $\leftarrow$ small, medium; good $\leftarrow$ small, large; good $\leftarrow$ medium, medium; scared $\leftarrow$ large, medium; scared $\leftarrow$large, large; kamikaze$\leftarrow$small, small; kamikaze$\leftarrow$ medium, small; bad $\leftarrow$ large, small; scared in any other case}.
    \end{itemize}

The aggregation function g used to determine the degrees of validity of the 2CP corresponds to an average of the degrees of validity of the 1CPs that determine their value.

\section{Classification}
\label{classification}

Agent learning was posed as a classification problem where the CPs associated with the actions of a human player are used as training data to implement a resulting static classifier in the video game to decide the movements of an agent according to its own CPs. A decision tree without pruning was used as a classification method. This method is very useful in this case since the resulting classifiers can easily be expressed as a set of if-then rules. In our case, J48 provides us with static classifiers that can be easily implemented in our video game. The attributes used for the generation of the classifiers were defined from CPs, and two attributes were assigned for each perception, one for the linguistic variable and another for the degree of validity. The tags used were constant throughout the tests. The class was defined as an action {go ahead, get away}.

The implementation of the static classifiers generated from these tests was performed using the code generated by the Weka library, which exports the source code as a class in Java. This can be implemented directly since the game is also written in Java. The data used for decision making with the static classifiers are CPs obtained in real time during a game.

\subsection{Tests performed}

Three tests are performed on decision trees changing the CPs used and the way in which the class values are determined in the training data set. 

\pgfkeys{
    /pgf/number format/precision=0, 
    /pgf/number format/fixed zerofill=true
}
\begin{figure}
\begin{center}
    \begin{tikzpicture}
        \centering
        \begin{axis}[
            ybar, axis on top,
            title={Number of attributes per test},
            height=6cm, width=6cm,
            bar width=0.4cm,
            ymajorgrids, tick align=inside,
            major grid style={
                draw=white,
                },
            enlarge y limits={value=.1,upper},
            ymin=0, ymax=20,
            axis x line*=bottom,
            axis y line*=left,
            y axis line style={opacity=0},
            tickwidth=0pt,
            enlarge x limits=true,
            legend style={
                at={(1.3,0.5)},
                anchor=north,
                legend columns=1,
                /tikz/every even column/.append style={
                    column sep=0.0cm,
                    anchor=west
                }
            },
            ylabel={Number of attributes},
            ytick={0,4,7,10,13,16,19},
            symbolic x coords={
                T1,T2,T3},
            xtick=data,
            nodes near coords={
                \pgfmathprintnumber[fixed]{\pgfplotspointmeta}
            }
        ]
            \addplot [draw=none, fill=blue!30] coordinates {
                (T1,7)
                (T2,9) 
                (T3,9) };
            \addplot [draw=none, fill=blue] coordinates {
                (T1,-1)
                (T2,17) 
                (T3,17) };
    
            \legend{First part,Second part}
        \end{axis}
    \end{tikzpicture}
\end{center}
\caption{Attribute comparison.}
\end{figure}
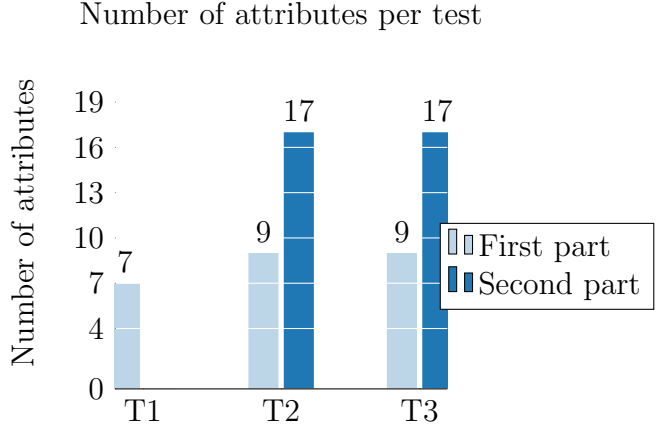

The first test (T1) points to the use of a minimum number of 1CP in the training data set. Three 1CPs were used, chosen based on the relevance of the metrics with which they were constructed. The 1CP used corresponded to the Protection CP, Distance CP (player-opponent), and Closeness CP. The latter was used exclusively for this test and is defined as follows.

\begin{itemize}
 \setlength\itemsep{1em}
 \item \textbf{CP Closeness.} This is the perception used to perceive how much closer or farther one entity is from another in relation to a third entity. Given a scenario divided into cells of equal size and considering that an entity occupies the space of a cell, the proximity is calculated according to the distance between entities, which is determined by the number of cells between them. We define Closeness C as the difference between the distances from entity a to entity b and from entity b to entity c. This can be calculated using the following equation: C = d (a, b) - d (b, c). In particular, we use this CP to perceive how much closer or farther the player is to the closest reward in relation to the opponent. We define this 1CP as follows:
Z = [−N; N], where N is the maximum distance in the game scenario.
A = ("much closer"; "closer"; "farther away"; "much farther away").
g: This function is constructed using four labels that are represented
with trapezoidal membership functions: "much closer" (2,6, N, N), "more
close" (- 2,2,2,6), "farther" (- 6, -2, -2,2), and "much farther" (- N, -N, -6, - 2). 
\end{itemize}

\begin{table}
\begin{tabular}{ |p{2.0cm}|p{3.0cm}|p{2.0cm}|  }
    \hline
        \tiny Attribute & \tiny Description & \tiny Values \\
    \hline
        \tiny CP Protection & 
        \tiny Number of obstacles between player and opponent. & 
        \tiny low, medium, high \\
    \hline
        \tiny DV Protection & \tiny Degrees of validity for CP Protection & \tiny Numeric (Real) \\
    \hline 
        \tiny CP Distance \newline (player-opponent) & \tiny Distance between player and opponent. & \tiny small, medium, large \\
    \hline
        \tiny DV Distance & \tiny Degrees of validity for CP Distance & \tiny Numeric (Real) \\
    \hline 
        \tiny CP Closeness & \tiny Perception used to perceive how much closer or farther one entity is from another in relation to a third entity & \tiny "much closer"; "closer"; "farther away"; "much farther away" \\
    \hline
        \tiny DV Closeness & \tiny Degrees of validity for CP Closeness. & \tiny Numeric (Real) \\
        \hline
    \tiny \textbf{Action} & \tiny \textbf{Action performed by the player (the class).} & \tiny \textbf{``go ahead'', ``get away'' } \\
    \hline 
\end{tabular}
\caption{Test 1 attribute list.}
\end{table}

In this test, the class label values ​​of each instance of the training dataset are determined by a simple rule based on the distance metric (player-opponent). The rule is that if the distance between the player and the opponent is maintained or increased, then the player is evading the opponent unless the player is ignoring the adversary. The second test (T2) points to the use of all the perceptions defined in the CP section as attributes keeping the class definition. The values of the class label of each instance of the training dataset are obtained in the same way as previously described. This test was divided into two parts: in the first part, a test was performed using all the CPs as attributes, and in the second part, another test was carried out using the 2CPs CP Situation, CP Attitude and CP Movement together with the 1CP Time as attributes.

\begin{table}
\begin{center}
\begin{tabular}{ |p{2.0cm}|p{3.0cm}|p{2.0cm}|  }
    \hline
        \tiny Attribute & \tiny Description & \tiny Values \\
    \hline 
        \tiny CP Distance \newline (opponent-reward) & \tiny Distance between opponent and closest  reward to the player & \tiny small, medium, large \\
    \hline
        \tiny DV Distance & \tiny Degrees of vality for CP Distancia. & \tiny Numeric (Real) \\
    \hline 
        \tiny CP Distance \newline (player-reward) & \tiny Distance between player and closest reward to the player. & \tiny small, medium, large \\
    \hline
        \tiny DV Distance & \tiny Degrees of vality for CP Distance. & \tiny Numeric (Real) \\
    \hline 
        \tiny CP Distance \newline (player-opponent) & \tiny Distance between player and opponent. & \tiny small, medium, large \\
    \hline
        \tiny DV Distance & \tiny Degrees of vality for CP Distance. & \tiny Numeric (Real) \\
    \hline
        \tiny CP Protection & 
        \tiny Number of obstacles between player and opponent. & 
        \tiny low, medium, large \\
    \hline
        \tiny DV Protection & \tiny Degrees of vality for CP Protection & \tiny Numeric (Real) \\
    \hline 
        \tiny CP Time & \tiny time elapsed since the start of the game & \tiny little, reasonable, alot \\
    \hline
        \tiny DV Time & \tiny Degrees of validy for CP Time & \tiny Numeric (Real) \\
    \hline 
        \tiny CP Attitude & \tiny current attitude of the player & \tiny wise, brave, prudent, passive \\
    \hline
        \tiny DV Attitude & \tiny Degrees of validy for CP Attitude & \tiny Numeric (Real) \\
    \hline 
        \tiny CP Movement & \tiny current movement perfomed by the player & \tiny good, scared, kamikaze, bad \\
    \hline
        \tiny DV Movement & \tiny Degrees of validy for CP Movement. & \tiny Numeric (Real) \\
    \hline 
        \tiny CP Situation & \tiny current situation between player and opponent & \tiny risky, dangerous, safe, easy \\
    \hline
        \tiny DV Situation & \tiny Degrees of validy for CP Situation. & \tiny Numeric (Real) \\
    \hline
        \tiny \textbf{Action} & \tiny \textbf{Action performed by the player (the class).} & \tiny \textbf{go ahead, get away } \\
    \hline 
\end{tabular}
\caption{Test 2 and 3 attribute list.}
\end{center}
\end{table}

The third test (T3), similar to the second, was divided into two parts with the same end, pointing to the use of all the perceptions defined in the section of CP as attributes maintaining the class definition, but with the difference that the values ​​of the class label of each instance of the set of training data is determined differently, with a new set of rules based on the 2CPs used as attributes: CP Situation, CP Attitude and CP Movement. The rules used are the following: R1: go\_ahead $\leftarrow$ wise, good, safe; R2: go\_ahead $\leftarrow$ wise, scared, safe; R3: go\_ahead $\leftarrow$ wise, good, easy; R4: go\_ahead $\leftarrow$ wise, scared, easy; R5: go\_ahead $\leftarrow$wise, scared, safe; R6: go\_ahead $\leftarrow$ wise, scared, easy; get\_away in any other case.

\begin{figure*}
\label{fig-results}
\begin{center}
\begin{tikzpicture}
    \centering
    \begin{axis}[
        ybar, axis on top,
        title={10-fold crossvalidation results},
        height=6cm, width=12cm,
        bar width=0.4cm,
        ymajorgrids, tick align=inside,
        major grid style={
            draw=white,
            },
        enlarge y limits={value=.1,upper},
        ymin=0.0, ymax=1.0,
        axis x line*=bottom,
        axis y line*=left,
        y axis line style={opacity=0},
        y tick label style={
            /pgf/number format/.cd,
                fixed,
                fixed zerofill,
                precision=1,
            /tikz/.cd
        },
        tickwidth=0pt,
        enlarge x limits=true,
        legend style={
            at={(0.5,-0.2)},
            anchor=north,
            legend columns=3,
            /tikz/every even column/.append style={
                column sep=0.25cm,
                anchor=west
            }
        },
        ylabel={},
        ytick={0,0.2,0.4,0.6,0.8,1.0},
        symbolic x coords={
            T1,T2(9),T2(17),T3(9),T3(17)},
        xtick=data,
        nodes near coords={
            \pgfmathprintnumber[fixed,precision=1]{\pgfplotspointmeta}
        }
    ]
        \addplot [draw=none, fill=blue!30] coordinates {
            (T1,0.910)
            (T2(9),0.876)
            (T2(17),0.890)
            (T3(9),1.000)
            (T3(17),1.000) };
        \addplot [draw=none, fill=blue] coordinates {
            (T1,0.914)
            (T2(9),0.858)
            (T2(17),0.912)
            (T3(9),1.000)
            (T3(17),1.000) };
        \addplot [draw=none, fill=blue!70!black] coordinates {
            (T1,0.932)
            (T2(9),0.887)
            (T2(17),0.921)
            (T3(9),1.000)
            (T3(17),1.000) };

        \legend{Accuracy,ROC Area,PRC Area}
    \end{axis}
\end{tikzpicture}
\caption{Bar graph comparing different classifiers (T1, T2(9), T2(17), T3(9), T3(17)) according to performance measures}
\end{center}
\end{figure*}
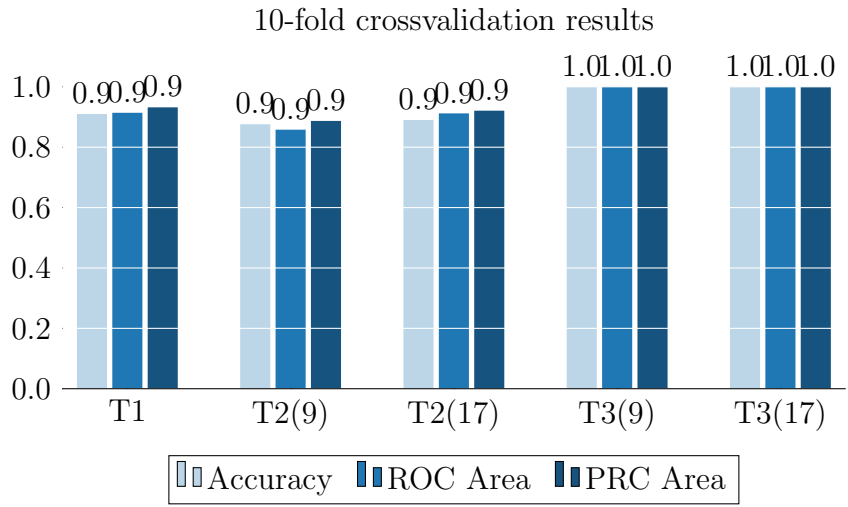

\section{Evaluation}

\begin{table*}
\begin{center}
\begin{tabular}{|p{1.8cm}|p{1.3cm}|p{1.3cm}|p{1.3cm}|p{1.3cm}|p{1.3cm}|}
    \hline
        & T1 & T2(9) & T2(17) & T2(9) & T2(17) \\
    \hline
        Accuracy & 0.910 & 0.876 & 0.890 & 1.000 & 1.000 \\
        ROC Area & 0.914 & 0.858 & 0.912 & 1.000 & 1.000 \\
        PRC Area & 0.932 & 0.887 & 0.921 & 1.000 & 1.000 \\
    \hline
\end{tabular}
\caption{Detailed Accuracy.}
\end{center}
\end{table*}

The tests were conducted with a dataset of approximately 1,160 instances. We evaluated the classifiers with the validation method of 10-fold cross validation. From the results of the validation, we evaluate the performance of the models based on the following metrics.

\begin{itemize}
\setlength\itemsep{1em}
\item \textbf{Accuracy:} We define accuracy as the proximity of a measured value and a reference value. In our case, we represent accuracy as the percentage of instances classified correctly. We use this metric to determine how correct the models are when classifying.
\item \textbf{ROC Area:} This area is obtained from the receiver operating characteristic curve, which is plotted in a graph of axes (1-specificity) and sensitivity. The area under the receiver operating characteristic curve is a value that represents the expected performance of a classifier considering the correctly and incorrectly classified instances. We use this metric to determine how well the models perform in predicting actions (evade or go ahead).
\item \textbf{PRC Area:} Obtained from the precision-recall curve, which is plotted on a precision and recall axis graph, the area under the precision-recall curve is a value representing the expected performance of a classifier considering correctly predicted instances and correctly classified instances. We use this metric to determine how well the models predict the actions considering an imbalance in the data.
\end{itemize}

Table 3 presents the main values ​​related to the evaluation of classifiers, from which it can be seen that the results were similar between the metrics of each test, showing that the models generated perform in a similar way through the metrics with which they are evaluated. From these results, we can conclude that the accuracy is representative of the real performance of the generated models and, therefore, that the generated models imitate the learned behaviour with a minimum accuracy close to 90\%. In the third test, the values ​​notoriously reach 100\%, which is due to how the class values ​​are obtained. By relying on rules that use the same values ​​of the attributes of the dataset, the rules of the generated classifier results from the inference of the rules that determine the values ​​of the class, ignoring any value irrelevant to the classification, as in this case would be the degrees of truth. Unfortunately, from this test, we can conclude that the training resulted in a classifier that mimics the rules that determine the value of the class rather than mimicking the behaviour of the player.

Through the observation of the agent's behaviour using the different classifiers, we found that, in general, the agent can finish the games successfully and without problems with almost all classifiers, except for the case of the classifier generated in the second test with 17 attributes (T2 (17)). This one showed a more inconsistent behaviour than the rest during the tests. On the other hand, a very similar classifier but with fewer rules, such as the result of the second test with 9 attributes (T2 (9)), was very effective. From this finding, it could be suggested that if it had been a pruned tree, the classifier with 17 attributes would have been much more effective. The classifier resulting from the first test was simple and effective, resulting in a set of rules smaller than those of the classifiers of the second test but much higher than those of the third test classifiers, since these turned out to have a minimum number of rules. In terms of the speed to successfully solve the video game scenario, the agent was progressively faster during the tests, starting with a game duration of approximately 2 minutes in the first test (T1) and ending near a minute and a half in the third test (T3). This is observable in the video demo files linked in the Introduction section.

\section{Conclusions}

Video games are a very powerful tool for investigating the techniques of artificial intelligence. We have pointed out the importance of the development of opponents in such environments and the limitations of the programming of their behaviour. In this sense, we have proposed a new way of programming behaviours by imitation, proposing and evaluating a machine learning model based on the linguistic description of complex phenomena and decision trees. The result was an architecture composed of three modules (perception, learning and action) that allow the creation of a correspondence between perceptions and movements. We have managed to make bots imitate human players based on the observations themselves, in our case, the games. From the results obtained after the evaluation, we can conclude that our model can be very useful for programming the behaviour of characters in video games since it uses fuzzy logic and possesses the ability to handle the imprecision and the underlying subjectivity in this process. 

In terms of future work, we can suggest that future studies focus on obtaining more effective classifiers for this purpose through pruned trees and the use of a random forest. We can also suggest the use of explainable artificial intelligence to obtain a better understanding and evaluation of learned behaviours.

\section*{acknowledgements}
This paper is the result of work by the SOMOS research group (SOftware – MOdelling – Science), funded by the Direcci\'on de Investigaci\'on and Facultad de Ciencias Empresariales of the Universidad del B\'io-B\'io, Chile. The authors thank the Facultad de Ingenier\'ia de la Universidad Cat\'olica de la Sant\'isima Concepci\'on, Chile.

\section*{Authors Contribution}

All authors have contributed equally to the work.
All authors contributed to the study conception,  design and implementation. Material preparation, data collection and analysis were performed by Clemente Rubio-Manzano and Tom\'as Lermanda. The first draft of the manuscript was written by Clemente Rubio-Manzano and Tom\'as Lermanda and all authors commented on previous versions of the manuscript. All authors read and approved the final manuscript.

\section*{Conflict of interest}

Clemente Rubio, Tom\'as Lermanda, Claudia Mart\'inez, Christian Vidal and Alejandra Segura declare that they have no conflict of interest.

%
%


\begin{thebibliography}{10}

\bibitem{conde2018toward}
Conde-Clemente, P., Alonso, J. M., and Trivino, G. 
\newblock Toward automatic generation of linguistic advice for saving energy at home. 
\newblock Soft Computing, 22(2), 345-359, 2018.

\bibitem{borovikov2019towards}
I.~Borovikov, J.~Harder, M.~Sadovsky, and A.~Beirami.
\newblock Towards interactive training of non-player characters in video games.
\newblock {\em arXiv preprint arXiv:1906.00535}, 2019.

\bibitem{feng2016towards}
S.~Feng and A.-H. Tan.
\newblock Towards autonomous behavior learning of non-player characters in
  games.
\newblock {\em Expert Systems with Applications}, 56:89--99, 2016.

\bibitem{fod2002automated}
A.~Fod, M.~J. Matari{\'c}, and O.~C. Jenkins.
\newblock Automated derivation of primitives for movement classification.
\newblock {\em Autonomous robots}, 12(1):39--54, 2002.

\bibitem{gemine2012imitative}
Q.~Gemine, F.~Safadi, R.~Fonteneau, and D.~Ernst.
\newblock Imitative learning for real-time strategy games.
\newblock In {\em 2012 IEEE Conference on Computational Intelligence and Games
  (CIG)}, pages 424--429. IEEE, 2012.

\bibitem{marcotte2017behavior}
R.~Marcotte and H.~J. Hamilton.
\newblock Behavior trees for modelling artificial intelligence in games: A
  tutorial.
\newblock {\em The Computer Games Journal}, 6(3):171--184, 2017.

\bibitem{melendez2009controlling}
P.~Melendez.
\newblock Controlling non-player characters using support vector machines.
\newblock In {\em Proceedings of the 2009 Conference on Future Play on@ GDC
  Canada}, pages 33--34, 2009.

\bibitem{Murthy1998}
S.~K. Murthy.
\newblock Automatic construction of decision trees from data: A
  multi-disciplinary survey.
\newblock {\em Data Mining and Knowledge Discovery}, 2(4):345--389, Dec 1998.

\bibitem{Witten2011}
T.~Ngo.
\newblock Data mining: Practical machine learning tools and technique, third
  edition by ian h. witten, eibe frank, mark a. hell.
\newblock {\em SIGSOFT Softw. Eng. Notes}, 36(5):51–52, Sept. 2011.

\bibitem{rubio2019fuzzy}
C.~Rubio-Manzano, T.~Lermanda~Senoceain, C.~Martinez-Araneda, C.~Vidal-Castro,
  and A.~Segura-Navarrete.
\newblock Fuzzy linguistic descriptions for execution trace comprehension and
  their application in an introductory course in artificial intelligence.
\newblock {\em Journal of Intelligent \& Fuzzy Systems}, 37(6):8397--8415,
  2019.

\bibitem{RUBIOMANZANO201627}
C.~Rubio-Manzano and G.~Trivino.
\newblock Improving player experience in computer games by using players'
  behavior analysis and linguistic descriptions.
\newblock {\em International Journal of Human-Computer Studies}, 95:27 -- 38,
  2016.

\bibitem{thurau2004imitation}
C.~Thurau, C.~Bauckhage, and G.~Sagerer.
\newblock Imitation learning at all levels of game-ai.
\newblock In {\em Proceedings of the international conference on computer
  games, artificial intelligence, design and education}, volume~5, 2004.

\bibitem{thurau2004learning}
C.~Thurau, C.~Bauckhage, and G.~Sagerer.
\newblock Learning human-like movement behavior for computer games.
\newblock In {\em Proc. Int. Conf. on the Simulation of Adaptive Behavior},
  pages 315--323, 2004.

\bibitem{thurau2007bayesian}
C.~Thurau, T.~Paczian, G.~Sagerer, and C.~Bauckhage.
\newblock Bayesian imitation learning in game characters.
\newblock {\em International journal of intelligent systems technologies and
  applications}, 2(2-3):284--295, 2007.

\bibitem{van1999learning}
M.~Van~Lent and J.~Laird.
\newblock Learning hierarchical performance knowledge by observation.
\newblock In {\em ICML}, pages 229--238, 1999.

\bibitem{zadeh2001new}
L.~A. Zadeh.
\newblock A new direction in ai: Toward a computational theory of perceptions.
\newblock {\em AI magazine}, 22(1):73--73, 2001.

\end{thebibliography}


\end{document}